\def\year{2022}\relax
\documentclass[letterpaper]{article} 
\usepackage{aaai22}  
\usepackage{times}  
\usepackage{helvet}  
\usepackage{courier}  
\usepackage[hyphens]{url}  
\usepackage{graphicx} 
\usepackage{natbib}  
\usepackage{caption} 
\DeclareCaptionStyle{ruled}{labelfont=normalfont,labelsep=colon,strut=off} 
\usepackage{algorithm}
\usepackage{algorithmic}
\usepackage{booktabs} 
\usepackage{amsfonts} 
\usepackage{multirow} %
\usepackage{amsmath}

\usepackage{graphicx}
\graphicspath{{./figures/}}
\usepackage{amsmath}
\usepackage{multirow}
\usepackage{enumerate}
\usepackage{arydshln}
\usepackage{stfloats}
\usepackage{booktabs}
\usepackage[flushleft]{threeparttable}

\usepackage{newfloat}
\usepackage{listings}
\floatstyle{ruled}
\newfloat{listing}{tb}{lst}{}
\floatname{listing}{Listing}
\title{Graph4Rec: A Universal Toolkit with Graph Neural Networks for Recommender Systems}
\title{Graph4Rec: A Universal Toolkit with Graph Neural Networks for Recommender Systems}
\author{
	Weibin Li,
	Mingkai He, 
	Zhengjie Huang, 
	Xianming Wang, \\
	Shikun Feng, 
	Zhihua Wu,
	Weiyue Su, 
	Yu Sun
}
\affiliations {
    Baidu Inc., China\\
	\{liweibin02, hemingkai, huangzhengjie, wangxianming, fengshikun01, wuzhihua02, suweiyue, sunyu02\}@baidu.com
}

\usepackage{bibentry}

\usepackage{hyperref}

\usepackage{xcolor}
\usepackage{lineno}
\usepackage{lipsum}

\begin{document}

\maketitle
\begin{abstract}
	In recent years, owing to the outstanding performance in graph representation learning, graph neural network (GNN) techniques have gained considerable interests in many real-world scenarios, such as recommender systems and social networks. In recommender systems, the main challenge is to learn the effective user/item representations from their interactions. However, many recent publications using GNNs for recommender systems cannot be directly compared, due to their difference on datasets and evaluation metrics. Furthermore, many of them only provide a demo to conduct experiments on small datasets, which is far away to be applied in real-world recommender systems. To address this problem, we introduce Graph4Rec, a universal toolkit that unifies the paradigm to train GNN models into the following parts: graphs input, random walk generation, ego graphs generation, pairs generation and GNNs selection. From this training pipeline, one can easily establish his own GNN model with a few configurations. Besides, we develop a large-scale graph engine and a parameter server to support distributed GNN training. We conduct a systematic and comprehensive experiment to compare the performance of different GNN models on several scenarios in different scale. Extensive experiments are demonstrated to identify the key components of GNNs. We also try to figure out how the sparse and dense parameters affect the performance of GNNs. Finally, we investigate methods including negative sampling, ego graph construction order, and warm start strategy to find a more effective and efficient GNNs practice on recommender systems. Our toolkit is based on PGL \footnote{https://github.com/PaddlePaddle/PGL} and the code is opened source in \href{https://github.com/PaddlePaddle/PGL/tree/graph4rec/apps/Graph4Rec}{https://github.com/PaddlePaddle/PGL/tree/main/apps/Graph4Rec}.
	
\end{abstract}

\section{Introduction}
Learning representation from a user-item graph is ubiquitous in modern deep learning-based Recommender Systems (RS), such as news recommendations and product suggestions on e-commerce websites. In these applications, the interactions between users and items construct a large-scale heterogeneous graph. The representation of nodes learned from the graph structure, i.e., the low dimension vectors of users and items can be applied to downstream applications. For example, user-based collaborative filtering methods can predict the likely-to-interact items for a user based on ratings given to that item by the other users with similar tastes, which are measured by the similarity of user embeddings. While for item-based collaborative filtering methods, item embeddings are leveraged for similar products recommendation. Therefore, learning meaningful representations from complex graph structures plays a vital role in the development of recommender systems. 

Over the past decade, a series of studies have been explored for mining homogeneous or heterogeneous graphs.  For example, walk-based algorithms generate node sequence by defined random walk strategies \cite{perozzi2014deepwalk,dong2017metapath2vec,wang2018billion}. Then they pull closer the nodes inside the same window and push away those from random sample nodes. Recently, as Graph Neural Networks (GNNs) have shown their capability of modeling complex structural data, 
many existing works attempt to adopt GNNs for representation learning and recommender systems. 
GNN models iteratively aggregate neighbors for node representations and utilize the high-order network information for improvements \cite{kipf2016semi,velivckovic2017graph,hamilton2017inductive,xu2018powerful}. Despite the research on GNNs' architecture, there are also many studies focusing on developing effective tasks for graph learning such as contrastive learning and self-supervised learning \cite{qiu2020gcc,you2020graph,wu2021self}. However, many recent works related to GNNs for recommender systems cannot be directly compared, due to their difference in datasets and evaluation metrics. 
Furthermore, many of them only provide a demo to conduct experiments on small datasets, which is far away from web-scale recommender systems in real world.

In recent years, there are several existed graph embedding systems,
such as GraphVite \cite{zhu2019graphvite} and PyTorch-BigGraph \cite{pbg} with PBG for short.
GraphVite only performs walk-based models on a single machine with multi-GPUs.
Although PBG support distributed training, it cannot deal with heterogeneous GNN models, 
lacking the capability of modeling complex structural data for recommender systems.
Besides, neither of the systems can handle side information of node in a graph to address the cold start problem.

In this paper, we introduce Graph4Rec, a universal toolkit that unifies the paradigm to train GNN models into the following parts: graphs input, random walk generation, ego graphs generation, pairs generation and GNNs selection. 
With a few configurations, researchers are freely to build their own GNN models.
Besides, unlike the traditional walk-based embedding system, the utilization of GNNs makes Graph4Rec more competitive while modeling complex user-item interactions data for recommender systems. The distributed graph engine and parameter server empower Graph4Rec to handle large-scale graph data. Furthermore, we provide systematic and comprehensive experiments to evaluate the performance of various GNNs in different scenarios and attempt to give out some suggestions for developing GNNs in the recommender system.

\section{Preliminaries}

In this section, we introduce the notation of heterogeneous graphs 
and briefly review some recent research
on GNN-based representation learning for recommender system.

\subsection{Heterogeneous Graph Structure}
Generally, the complex interactions between users and items in recommender systems can be simply regarded as a heterogeneous graph denoted as $G=(\mathcal{V}, \mathcal{E}, \mathcal{R})$, where $\mathcal{V}$ denotes the nodes, $\mathcal{E}$ represents edges, and $\mathcal{R}$ is the edge types. In recommender systems, $\mathcal{V}$ can be the set of users and items 
while $\mathcal{R}$ representing different relations between users and items, such as click and purchase. If we only have single type of relation between vertices, the heterogeneous graph will degenerate into homogeneous graph.

\subsection{Graph Neural Networks} 
Recent developments of GNNs show their strong capability for mining graph data \cite{kipf2016semi,hamilton2017inductive,velivckovic2017graph,xu2018powerful}. The main idea of GNNs is to iteratively update the representation of nodes by aggregating representations of its neighbors. And most of them follow message passing paradigm \cite{gilmer2017neural}. After $K$ iterations of aggregation, a node’s representation captures the structural information within its K-hop neighborhood. Formally, the calculation in the $k$-th layer of a GNN is given as follows:
\begin{equation}
	\begin{split}
		\hat{h}_{v}^{k} &= \mathrm{AGGREGATE}^{k} \left( \left \{ h_{u}^{k-1}: u \in \mathcal{N}_v \right \} \right) \\
		h_{v}^{k} &= \mathrm{COMBINE}^{k} \left (h_v^{k-1}, \hat{h}_{v}^{k} \right )
	\end{split},
	\label{equ:message_passing_paradigm}
\end{equation} 
where $h_{v}^{k}$ is the representation of node $v$ at the $k$-th layer. And $\mathcal{N}_v$ stands for the neighborhood set of $v$. The $\mathrm{AGGREGATE}$ function can be implemented by a number of order-independent operations, 
such as $max$, $mean$, and $sum$. The $\mathrm{COMBINE}$ function is usually a simple neural network layer with feature transformation and non-linear activation. And neighborhood sampling is the most common way to address the exponential growth problem caused by K-hop neighbors \cite{hamilton2017inductive}.

\subsection{Graph Representation Learning}
The performance of the recommender systems usually depends on the quality of the user and item representation learnt from the interaction data. And the embeddings are often used in measuring the similarity between users and items for user-based or item-based collaborative filtering algorithms. For example, Wang \shortcite{wang2018billion} constructs item-item graph from the user log and learns representation for items. Then it recommends similar items according to user history.
Inspired by word2vec \cite{mikolov2013distributed}, early attempts on graph representations learning are skip-gram based models with pre-defined random walks on graph \cite{perozzi2014deepwalk,dong2017metapath2vec,wang2018billion}. Recent approaches leverage high-order information from graph structure with GNNs rather than simple ID embeddings. Although the unsupervised graph representation learning methods are named differently as contrastive learning with augmentation \cite{qiu2020gcc,you2020graph} or self-supervised learning \cite{wu2021self}, they both follow the same idea that pull closer the positive pairs with similar structure and push away from negatives. The loss function can be formulated as follows:
\begin{equation} 
	\mathcal{L}= - \log\ \sigma(y_{vu}) - \sum_{m=1}^{M} \mathbb{E}_{w_m \sim P(w)}[\log\ \sigma(- y_{w_mu})],
\end{equation}
where $y_{vu}$ takes the inner product  $h_v^{T}h_u$ from the final node representations, $\sigma(z) = \frac{1}{1+e^{-z}}$ and $P(w)$ is the distribution from which a negative node $w_m$ is drew from for $M$ times.

\section{Framework}
\begin{figure*}[htbp]
	\centering
	\includegraphics[scale=0.5]{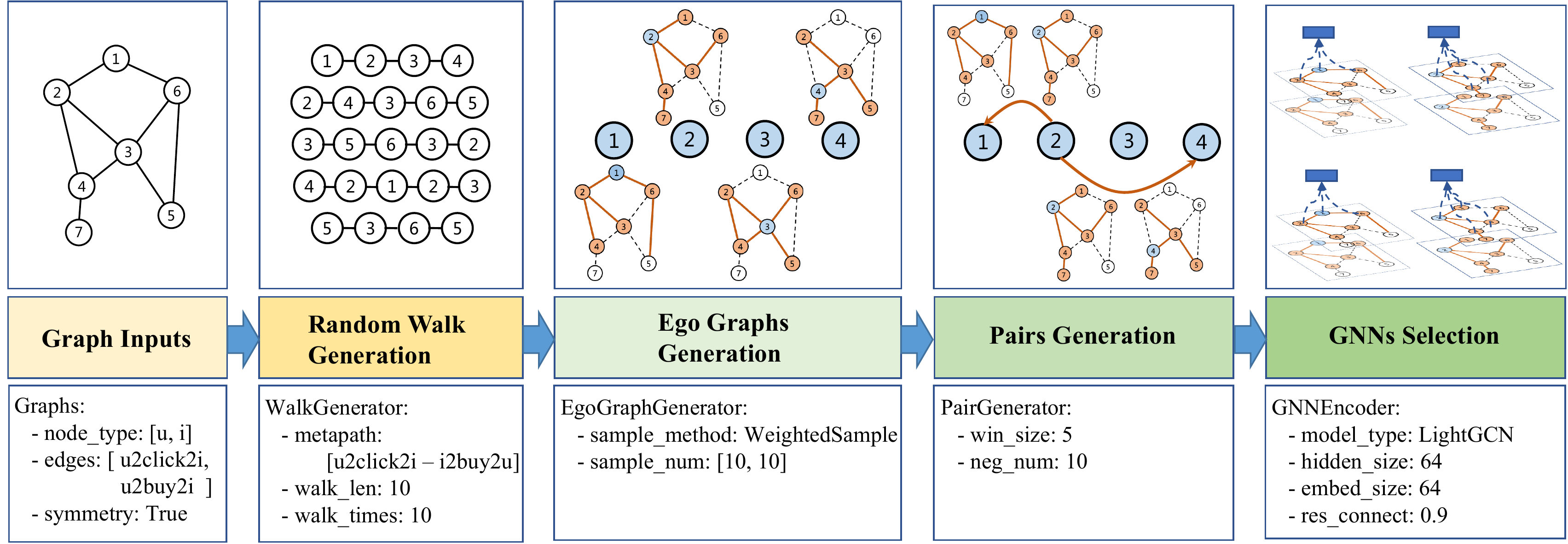}
	\caption{Graph4Rec unifies the paradigm to train GNN models 
		into the following parts: graphs input, random walk generation, 
		ego graphs generation, pairs generation and GNNs selection.}
	\label{fig:architecture}
	\vspace{-0.3cm}
\end{figure*}

As shown in Figure \ref{fig:architecture}, the paradigm of GNNs training process for recommender system consists of the following five components: graphs input, random walk generation, ego graphs generation, pairs generation, and GNNs selection. Multiple configuration settings in each component are provided to better control the graph embedding system flexibly. In the following sections, we describe the framework and the implementation of each component in Graph4Rec in detail. 
\subsection{Heterogeneous Graph Settings}
\label{label:graph_inputs}
In Graph4Rec, the basic data structure is the heterogeneous graph, 
in which nodes and edges have multiple types. A heterogeneous graph can be decomposed into several bipartite-directed graphs in which a triple $(u, r, v)$ is denoted as the source node, relation, and destination node. This general design of graph inputs helps us easily handle complex graph data. For example, in recommender systems, 
there are two types of nodes, i.e., user node and item node, represented by ``u'' and ``i''.
And the ``click'' interaction between ``u'' and ``i'' can be denoted as ``u2click2i''. For convenience, we use ``2'' as a delimiter to split the string into a triple, and the reverse relation of ``i2click2u'' will be automatically added when symmetry is true.  
If we only have a single type of nodes and edges, the heterogeneous graph will degenerate into a homogeneous graph, in which we can set the node type to ``u'' and the edge type to ``u2u''.

\subsubsection{Distributed Graph Engine} Meanwhile, Graph4Rec designs a distributed graph engine to deal with large-scale heterogeneous graph data. Nodes are partitioned uniformly into several machines. And the adjacency list of each node is stored in the corresponding server. To lower the communication cost between processes and machines, we optimize the data generation pipeline which will be discussed later.

\subsection{Random Walk Generation} 
Defining proximity using nodes within the same random path is the most essential method for graph representation learning \cite{perozzi2014deepwalk,qiu2020gcc}. As for heterogeneous graph mining,  meta path random walk \cite{dong2017metapath2vec} is adopted as a basic operation in Graph4Rec with its completeness satisfying most situations. Metapath can be simply defined as shown in Figure \ref{fig:architecture} that a sequence of edge types assembled head-to-tail with a hyphen. Inspired by the metapath2vec model, we develop a multi-metapaths random walk strategy that can sample multiple meta-paths from the heterogeneous graph. 
For example, given a heterogeneous graph described in Section \ref{label:graph_inputs}, 
we can specify the two metapaths, ``u2click2i - i2click2u'' and ``u2buy2i - i2buy2u'', to generate different types of paths to learn more structure information from the graph.
As for the homogeneous graph, we can set the metapath to ``u2u - u2u'', which is equal to a random walk. 
\subsection{Ego Graphs Generation}
For every node in the training samples, neighborhood sampling is required to reduce the computation cost for the later multi-hop neighbor aggregation in GNNs. In this work, we use an ego graph to represent a training sample of a central node. The definition of ego graph is that its node is composed of a central node and its neighbors.
For a node $v$, its neighbors of type $r$ are defined as $S_{v,r} = \{u: d(u, r, v) \leq K\}$,
where $d(u,r,v)$ is the shortest path distance between $u$ and $v$ of type $r$.
Thus, an ego graph of node $v$ in type $r$, denoted as $G_{v,r}$, is the subgraph induced by $S_{v,r}$.

Since there are multiple edge types, 
we then develop a relation-wise neighbor sampling method to 
allow relation-wise aggregation as described in Section \ref{label:aggregation}.
Formally, a relation-wise ego graph is denoted as $G_{v}$,
where $G_{v} = \{G_{v,r}: r \in \mathcal{R} \}$. Therefore, each node inside the path received from random walk generation becomes a central node. And thus nodes in the same paths batch will form their disjoint ego graphs with relation-wise neighborhood sampling, as shown in Figure \ref{fig:architecture}. Besides, ego graphs generation can be skipped if one only wants to train a walk-based model.

\subsection{Pairs Generation}
Pairs inside a random path are usually used to define proximity for contrastive learning or self-supervised learning on graphs \cite{qiu2020gcc,wu2021self}. Besides, items or users in the same path like ``u-i-u-i'' can be a potential recommendation result. Because the path implies that the users have the same behaviors, and the items have the same user groups. In this component, as shown in \ref{fig:architecture}, win\_size is used for the user to control the definition of proximity in a meta path. Then we generate ego graph pairs as positive training samples for the next procedure.

\subsection{GNNs Selection}
\label{label:aggregation}

After getting pairs of training samples, GNNs are then selected for ego graph encoding. For all the heterogeneous settings, we follow the idea from R-GCN \cite{RGCN} and provide neighborhood aggregation with distinct weights for each individual relation type. The node representations are calculated via
\begin{equation}
	\begin{split}
		h_{v,r}^{k} &= \text{GNN}_r(h_{v}^{k-1}, \{h_{u}^{k -1}: u\in \mathcal{N}_{v,r}\})\\
		h_{v}^{k} &= \alpha h_{v}^{0} + (1 - \alpha) \sum_{r}^{R} \phi_{r}h_{v,r}^{k}
	\end{split}.
\end{equation}
$\text{GNN}_r$ can be any graph neural network model that follows the message passing paradigm defined by Eq. \ref{equ:message_passing_paradigm}. For convenience, Graph4Rec has embraced various classical GNNs, such as GCN, GAT, LightGCN, etc. Users can also develop their substitution models. 

Unlike homogeneous graph with single relation aggregation result,
$\phi_{r}$ denotes the weight for relation-wise aggregation. The simplest way is to keep it as a constant uniform with $\phi_{r}=1$ between relations. We also provide a learnable relation aggregation setting like GATNE \cite{gatne}, which adopt a shallow network to provide attention score for each relation calculated by $\phi_{r}=\text{softmax}(\mathbf{w}^{T}\text{tanh}(\mathbf{W}h_{v,r}))$.

Over-smoothing \cite{li2019deepgcns,chen2020simple} is the key drawback of vanilla GNNs. $\alpha$ is the control of the residual connection from the bottom features $h^{0}$ to the output to tackle this problem. It can also be regarded as feature propagation with Personal PageRank \cite{klicpera2018predict}, which is a simple but effective strategy.

To address the cold start issue, side information such as category, brand, or user profile can be utilized as a basic feature and combined with ID embedding. In Graph4Rec, we support configurable sparse features with multiple slots. And each slot can have multiple values to support features with a variable length such as text or tags.

In the rest of our paper, although we adopt the model's name in their original paper, the relation-wise operation and tricks for handling over-smoothing problems are applied to all the models for a fair comparison.

\subsection{Training Optimizations}
\label{label:training}

\subsubsection{Parameter Server.}
With the increasing scale of graph data, 
the capacity of sparse learnable parameters (IDs or side information) also grow rapidly, 
limiting the GNN models to be applied in large-scale recommender systems.
In this work, we adopt a parameter server, 
containing a key-value store of embeddings based on PaddlePaddle\footnote{https://github.com/PaddlePaddle/Paddle}.
At each training step, the embeddings are pulled remotely from the parameter server. Then the calculated gradients are pushed to the server for an asynchronous update. The lazy initialization occurs when the embedding is required for the first time, which is memory efficient.
Empowered by distributed graph engine and parameter server,
Graph4Rec can easily train a GNN model with large-scale graph data for recommender systems.

\subsubsection{Walk, Sample, Pair:  Order Matters.}
\label{label:order}
The order of the training sample generation is important to the speed of training GNNs.
The most intuitive generation order is that we first sample a path, and then for a current vertice inside the path, another vertice is selected within the same window to construct pair.  Then we sample the ego graph for each pair to feed into GNN models. However, this kind of method will produce many repetitive nodes and each of them must sample an ego graph, which leads to increased communication for graph engine.
To alleviate this problem, 
we exchange the order of pair generation and ego graph sampling.
That is, after generating a walk path, we first perform ego graph sampling for each vertice within the path and
then construct training pair for GNN models. Formally, suppose we have a sampled path of length $L$ and window size is $w$.
Then we will have $O(wL)$ ego graph sampling operations
when generating pairs before sampling ego graphs. But if we exchange the order, the sampling times can reduce to $O(L)$. 
Since the diversity of training samples has been reduced, in the later section we will conduct experiment to show the trade-off between speed and performance.

\subsubsection{In-batch Negative Sampling.}
The main idea of our framework is to learn the representation of each node, 
which can pull similar points (positive pairs) together 
while pushing away dissimilar points (negative pairs). 
Here, the positive pairs can be the observed interactions 
while the negative pairs are randomly selected from $\mathcal{V}$. 
However, random selection of negative samples is time consuming, 
especially in distributed training mode 
where nodes and their side information are saved in different machines. 
Therefore, we implement an improved version that uses in-batch negative sampling. 
We maximize the scores for linked nodes while minimizing the scores of other nodes in a batch. 
This method can reduce additional data input, thereby increasing the training speed.

\subsubsection{Pre-training and Parameters Warm Start.}
The traditional walk-based models are fast and effective. We first pre-train the sparse embeddings from walk-based models. Then we train a GNN-based model and inherit the parameters for fast convergence and performance improvement.

\section{Experiments}

In this section, we report the settings and results of our experiments on four public datasets, and study the following research questions: 
\begin{itemize}
	\item RQ1: Does our proposed Graph4Rec outperform other existing graph learning systems in both speed and performance?
	\item RQ2: What is the best practice performance of current graph representation learning? 
	\item RQ3: How does the modeling of side information affect the performance of GNN models?
	\item RQ4: What is the impact of in-batch negative sampling?
	\item RQ5: How does the order of sampling ego graph influence the training speed?
	\item RQ6: What is the relationship of convergence between walk-based models and GNN-based models?
\end{itemize}

\subsection{Datasets}

\begin{table}[htbp]
	\centering
	\resizebox{1.0\linewidth}{!}{  
		\begin{tabular}{lllll}
			\toprule
			Dataset           & RetailRocket  & Rec15     & Tmall     & UB         \\ \hline
			\#Users           & 38,317  & 506,495   & 199,314   & 969,529    \\
			\#Items           & 54,937  & 39,656    & 464,995   & 4,158,142  \\
			\#Clicks          & 335,478 & 5,348,701 & 6,561,486 & 67,880,635 \\
			\#Purchases       & 11,082  & 204,346   & 398,036   & 1,336,173  \\
			\#Carts     & 27,157  & -         & 2,801     & 3,739,151  \\
			\#Favorites & -       & -         & 443,699   & 2,076,940  \\
			\#Train           & 373,717 & 5,553,047 & 7,406,022 & 75,032,899 \\
			\#Val             & 19,961  & 639,286   & 766,622   & 8,502,866  \\
			\#Test            & 19,871  & 637,830   & 754,203   & 8,498,985  \\ 
			\bottomrule
		\end{tabular}
	}
	\caption{Statistics of the processed datasets used in our experiments.}
	\vspace{-0.2cm}
	\label{tab:datasets}
\end{table}
We conduct experiments on four publicly available heterogeneous datasets, 
i.e., 
RetailRocket\footnote{https://www.kaggle.com/retailrocket/ecommerce-dataset}, Rec15\footnote{https://recsys.acm.org/recsys15/challenge/}, 
Tmall\footnote{https://tianchi.aliyun.com/dataset/dataDetail?dataId=42} and 
UB\footnote{https://tianchi.aliyun.com/dataset/dataDetail?dataId=649}. 
RetailRocket includes more than 4 months of examining, adding-to-carts, and 
purchasing records in an e-commerce platform. 
Rec15 is a competition dataset published in RecSys Challenge 2015.
Tmall is another dataset released by IJCAI Competition 2015, 
which contains four common behaviors in e-commerce websites, 
including click, purchase, ad-to-cart and add-to-favorite. 
UB is a user behavior dataset collected by real e-commerce websites.

For all datasets, we remove users in which the number of interactions is smaller than 10 (5 for RetailRocket). 
For the RetailRocket dataset, we drop the later duplicated (user, item, behavior) triple and 
discard items that are interacted fewer than 5 times.
We sort the historical interactions of each user in the order of timestamp, 
and select the first 80\% of each sequence as the training set, 
the next 10\% as the valid set, and the rest as the test set. 
The statistics of the four processed datasets is shown in Table~\ref{tab:datasets}.
We will release the preprocessed dataset together with our toolkit.

\subsection{Recall Strategies and Evaluation Metrics}
In our experiment, there are three common recall strategies 
i.e., user-based collaborative filtering (UCF), item-based collaborative filtering (ICF) 
and U2I that recall items for each user. 
Specifically, the ICF strategy recalls the most similar top-N (N=$20$ for our experiment) items 
for each interacted item $i$ of user $u$ and recommends the top-K items 
that appear most frequently in the recalled item set. 
UCF strategy recalls the most similar top-N (N=$20$ for our experiment) users $u’$ for each user $u$, 
and recommends the top-$K$ frequent items by aggregating the interacted items of $u’$. 
The U2I strategy directly uses user embedding to retrieve item embedding, 
and recommend the most similar top-K item.

We evaluate the top-K recommendation performance via a common metric, i.e., recall, 
which measures how many items in our recommendation list actually hit the user’s interest. 
We use ICF@K, UCF@K and U2I@K to represent the recall indicator of a top-K recommendation list 
obtained by ICF strategy, UCF strategy and U2I strategy, respectively.

\subsection{Compared Systems}

\begin{table}[htbp]
	\centering
	\setlength\tabcolsep{3pt}
	\resizebox{0.9\linewidth}{!}{
		\begin{tabular}{c|c|c|c|c|c|c}
			\hline
			& \begin{tabular}[c]{@{}c@{}}Random \\ Walk\end{tabular} & \begin{tabular}[c]{@{}c@{}}Homo.\\ Graph\end{tabular} & \begin{tabular}[c]{@{}c@{}}Hetero.\\ Graph\end{tabular} & \begin{tabular}[c]{@{}c@{}}Side \\ Info.\end{tabular} & \begin{tabular}[c]{@{}c@{}}Distributed\\ Training\end{tabular} & \begin{tabular}[c]{@{}c@{}}Parameters\\ Warm  Start\end{tabular} \\ \hline
			PBG       &                                                        & $\checkmark$                                          &                                                         &                                                       & $\checkmark$                                                   &                                                                    \\
			GraphVite & $\checkmark$                                           & $\checkmark$                                          &                                                         &                                                       &                                                    & $\checkmark$                                                       \\
			Graph4Rec & $\checkmark$                                           & $\checkmark$                                          & $\checkmark$                                            & $\checkmark$                                          & $\checkmark$                                                   & $\checkmark$                                                       \\ \hline
		\end{tabular}
	}
	\caption{Features supported by each systems.}
	\label{table:support}
\end{table}

In order to study whether our Graph4Rec is competitive to the other existing graph learning systems, 
we select the below systems as baselines for comparison, 
and report the best performance of the algorithms supported by each system on the four datasets.

\begin{itemize}
	\item PyTorch-BigGraph (PBG) \cite{pbg}: 
	A distributed system for learning graph embeddings for large graphs, 
	particularly big web interaction graphs with up to billions of entities and trillions of edges.
	
	\item GraphVite \cite{zhu2019graphvite}: 
	A general graph embedding engine, dedicated to high-speed and large-scale embedding learning in various applications.
\end{itemize}

A detailed comparison of the above systems is shown in Table \ref{table:support}. 
GraphVite only performs walk-based models on a single machine.
PBG  supports training both on single and distributed environment, 
but it is particularly designed to learn multi-relation embedding for knowledge graphs
and neither of them can handle heterogeneous graphs and side information.
In contrast, our Graph4Rec has more comprehensive functions, 
which can process more complex structure data and models in recommender systems.

\subsection{Experimental Results}

\subsubsection{Performance on Different Systems (RQ1).}
We first evaluate our Graph4Rec on the four public recommender datasets compared with other three systems.
Since the three existing systems do not support GNN-based models, 
we then use the DeepWalk \cite{perozzi2014deepwalk} model (DistMult \cite{DistMult} for PBG) for comparison. 
For each model, we train it with 2 billion pairs (edges) and report the results.

\begin{table*}[htbp]
	\resizebox{1.0\linewidth}{!}{
		\begin{threeparttable}
			\setlength\tabcolsep{3pt}
			\begin{tabular}{lcccccccccccc}
				\hline
				\textbf{}                                                             & \multicolumn{3}{c}{\textbf{RetailRocket}}                                  & \multicolumn{3}{c}{\textbf{Rec15}}                                         & \multicolumn{3}{c}{\textbf{Tmall}}                                       & \multicolumn{3}{c}{\textbf{UB}}                     \\ \cline{2-13} 
				& \textbf{ICF@100} & \textbf{UCF@100} & \textbf{U2I@100}                     & \textbf{ICF@100} & \textbf{UCF@100} & \textbf{U2I@100}                     & \textbf{ICF@1k} & \textbf{UCF@1k} & \textbf{U2I@lk}                      & \textbf{ICF@1k} & \textbf{UCF@1k} & \textbf{U2I@1k} \\ \hline
				\multicolumn{1}{l|}{\textbf{DistMult (PBG )\tnote{\textdagger}}}      & 0.0987           & 0.1106           & \multicolumn{1}{c|}{0.0673}          & 0.3830           & 0.4549           & \multicolumn{1}{c|}{0.3258}          & 0.1480          & 0.0829          & \multicolumn{1}{c|}{0.1270}          & 0.1122          & 0.0415          & 0.0441          \\
				\multicolumn{1}{l|}{\textbf{DeepWalk (GraphVite)\tnote{\textdagger}}} & 0.1343           & 0.1551           & \multicolumn{1}{c|}{0.1034}          & \textbf{0.4809}  & 0.4448           & \multicolumn{1}{c|}{0.3504}          & 0.1765          & 0.1351          & \multicolumn{1}{c|}{0.1281}          & \textbf{0.1304} & \textbf{0.0769}          & 0.0756          \\ \hline
				\multicolumn{1}{l|}{\textbf{DeepWalk (ours)}}                         & 0.1314           & 0.1639           & \multicolumn{1}{c|}{0.1316}          & 0.4527           & 0.4768           & \multicolumn{1}{c|}{0.4348}          & 0.1643          & 0.1162          & \multicolumn{1}{c|}{0.1398}          & 0.1028          & 0.0638          & 0.0519          \\
				\multicolumn{1}{l|}{\textbf{metapath2vec (ours)}}                     & 0.1392           & 0.1650           & \multicolumn{1}{c|}{0.1354}          & 0.4563           & \textbf{0.4828}  & \multicolumn{1}{c|}{0.4362}          & 0.1654          & 0.1165          & \multicolumn{1}{c|}{0.1407}          & 0.1021          & 0.0638          & 0.0523          \\
				\multicolumn{1}{l|}{\textbf{LightGCN (ours)}}                         & \textbf{0.1625}  & \textbf{0.1684}  & \multicolumn{1}{c|}{\textbf{0.1602}} & 0.4517           & 0.4730           & \multicolumn{1}{c|}{\textbf{0.4473}}          & \textbf{0.1870} & \textbf{0.1361} & \multicolumn{1}{c|}{\textbf{0.1752}} & 0.1252          & 0.0751          & \textbf{0.0774} \\ \hline
			\end{tabular}
			\begin{tablenotes}
				\item[\textdagger]  Results obtained running code provided by the original authors. 
			\end{tablenotes}
		\end{threeparttable}
	}
	\vspace{-0.1cm}
	\caption{The performance of different models implemented with our Graph4Rec and other existing systems on four datasets.}
	\label{table:comp_systems}
	\vspace{-0.0cm}
\end{table*}

As shown in Table \ref{table:comp_systems}, 
the performance of DeepWalk model implemented by our Graph4Rec is competitive with others, 
verifying the capability of our Graph4Rec.
Moreover, we also present the performance of LightGCN (GNN-based model) \cite{he2020lightgcn} 
and find that it can achieve a better performance. 

\begin{figure}[htbp]
	\centering
	\includegraphics[width=\linewidth]{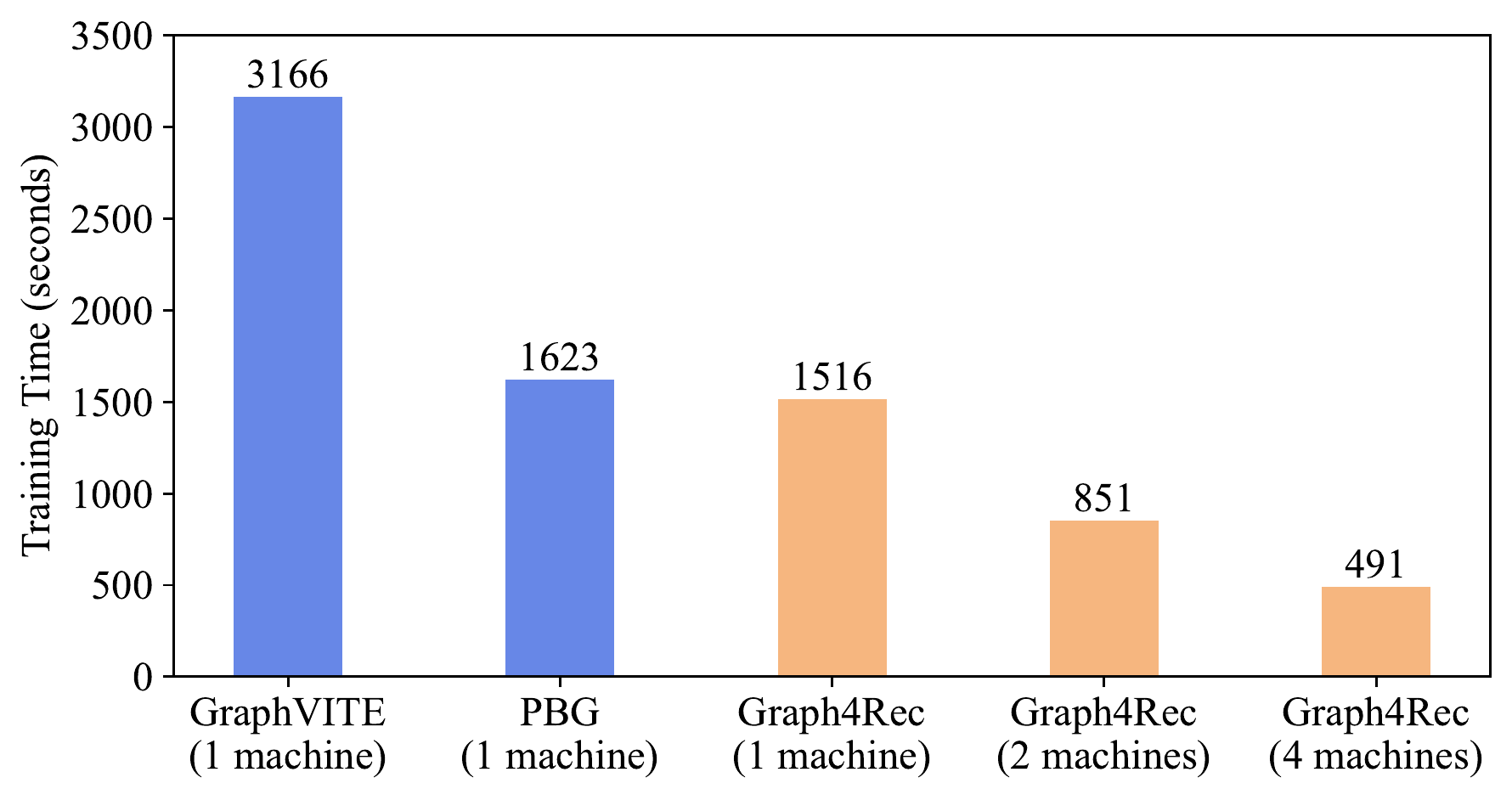}
	\vspace{-0.5cm}
	\caption{The runtime on Rec15 with 2 billion pairs.}
	\label{fig:speed}
\end{figure}

We further report the training speed of each existing systems on Rec15 dataset as shown in Figure \ref{fig:speed}. 
For fair comparison, we train the model on each system with 2 billion pairs (edges) on a 40-core CPU machine.
Since PBG do not contain the DeepWalk model, we use the DistMult \cite{DistMult} model instead.
Figure \ref{fig:speed} shows that, compared to GraphVite and PBG, our Graph4Rec can run faster, 
and even is 2$\times$ faster than GraphVite on a single machine. 
One reason for this is that 
we have done some optimizations for graph sampling, as decribed in Section \ref{label:training}.
Moreover, our Graph4Rec can linearly speed up the training process with distributed setting using multiple machines. 

\subsubsection{Performance in Graph4Rec (RQ2).}

\begin{table*}[htbp]
	\centering
	\resizebox{1.0\linewidth}{!}{
		\begin{threeparttable}
			\setlength\tabcolsep{3pt}
			\begin{tabular}{lcccccccccccc}
				\hline
				\textbf{}                                      & \multicolumn{3}{c}{\textbf{RetailRocket}}                                  & \multicolumn{3}{c}{\textbf{Rec15}}                                         & \multicolumn{3}{c}{\textbf{Tmall}}                                       & \multicolumn{3}{c}{\textbf{UB}}                     \\ \cline{2-13} 
				\textbf{}                                      & \textbf{ICF@100} & \textbf{UCF@100} & \textbf{U2I@100}                     & \textbf{ICF@100} & \textbf{UCF@100} & \textbf{U2I@100}                     & \textbf{ICF@1k} & \textbf{UCF@1k} & \textbf{U2I@1k}                      & \textbf{ICF@1k} & \textbf{UCF@1k} & \textbf{U2I@1k} \\ \hline
				\multicolumn{1}{l|}{\textbf{DeepWalk}}         & 0.1314           & 0.1639           & \multicolumn{1}{c|}{0.1316}          & 0.4527           & 0.4768           & \multicolumn{1}{c|}{0.4348}          & 0.1643          & 0.1162          & \multicolumn{1}{c|}{0.1398}          & 0.1028          & 0.0638          & 0.0519          \\
				\multicolumn{1}{l|}{\textbf{metapath2vec}}     & 0.1392           & \textbf{0.1650}  & \multicolumn{1}{c|}{0.1354}          & 0.4563           & 0.4828           & \multicolumn{1}{c|}{0.4362}          & 0.1654          & 0.1165          & \multicolumn{1}{c|}{0.1407}          & 0.1021          & 0.0638          & 0.0523          \\ \hline
				\multicolumn{1}{l|}{\textbf{GraphSAGE (mean)}} & 0.1369           & 0.1598           & \multicolumn{1}{c|}{0.1337}          & 0.4575           & 0.4774           & \multicolumn{1}{c|}{0.4448}          & 0.1708          & 0.1242          & \multicolumn{1}{c|}{0.1420}          & 0.1050          & 0.0659          & 0.0490          \\
				\multicolumn{1}{l|}{\textbf{GraphSAGE (sum)}}  & 0.1349           & 0.1561           & \multicolumn{1}{c|}{0.1299}          & 0.4547           & 0.4705           & \multicolumn{1}{c|}{0.4474}          & 0.1677          & 0.1238          & \multicolumn{1}{c|}{0.1496}          & 0.1002          & 0.0652          & 0.0528          \\
				\multicolumn{1}{l|}{\textbf{LightGCN}}         & \textbf{0.1543}  & 0.1597           & \multicolumn{1}{c|}{\textbf{0.1451}} & \textbf{0.4666}  & \textbf{0.4834}  & \multicolumn{1}{c|}{0.4439}          & \textbf{0.1848} & \textbf{0.1404} & \multicolumn{1}{c|}{0.1705}          & \textbf{0.1232} & \textbf{0.0770} & \textbf{0.0825} \\
				\multicolumn{1}{l|}{\textbf{GAT}}              & 0.1347           & 0.1589           & \multicolumn{1}{c|}{0.1273}          & 0.4617           & 0.4743           & \multicolumn{1}{c|}{\textbf{0.4562}} & 0.1680          & 0.1230          & \multicolumn{1}{c|}{0.1410}          & 0.1071          & 0.0672          & 0.0597          \\
				\multicolumn{1}{l|}{\textbf{GIN}}              & 0.1529           & 0.1587           & \multicolumn{1}{c|}{0.1210}          & 0.4546           & 0.4568           & \multicolumn{1}{c|}{0.4259}          & 0.1829          & 0.1371          & \multicolumn{1}{c|}{0.1593}          & 0.1205          & 0.0710          & 0.0721          \\
				\multicolumn{1}{l|}{\textbf{NGCF}}             & 0.1239           & 0.1520           & \multicolumn{1}{c|}{0.1063}          & 0.4503           & 0.4765           & \multicolumn{1}{c|}{0.4464}          & 0.1620          & 0.1154          & \multicolumn{1}{c|}{0.1331}          & 0.0983          & 0.0621          & 0.0463          \\
				\multicolumn{1}{l|}{\textbf{GATNE}}            & 0.1387           & 0.1609           & \multicolumn{1}{c|}{0.1342}          & 0.4601           & 0.4817           & \multicolumn{1}{c|}{0.4321}          & 0.1826          & 0.1359          & \multicolumn{1}{c|}{\textbf{0.1708}} & 0.1142          & 0.0713          & 0.0678          \\ \hline
			\end{tabular}

		\end{threeparttable}
	}
	\caption{The performance of different GNN models implemented by our Graph4Rec.}
	\label{table:performance}
\end{table*}

We also conduct a series of experiments to find out 
what is the best practice performance of current graph representation learning for recommender systems.
As shown in Table \ref{table:performance}, metapath2vec, i.e., heterogeneous random walk,  
achieves better result than DeepWalk, i.e., homogeneous random walk, which means that heterogeneous graph structure plays a significant role in representation learning.

To further verify the ability of different GNN models in our Graph4Rec toolkit, 
we implement some GNN models including GraphSAGE, LightGCN, GAT, GIN, NGCF and GATNE.
In general, LightGCN can outperform other GNN-based models as shown in Table \ref{table:performance}.
The results of this experiments verify the description of the paper of LightGCN \cite{he2020lightgcn}, 
that is, in large-scale graph representation learning, 
linear transformation has no positive effect on the effectiveness of collaborative filtering.

\begin{figure*}[htbp]
	\centering
	\includegraphics[width=\linewidth]{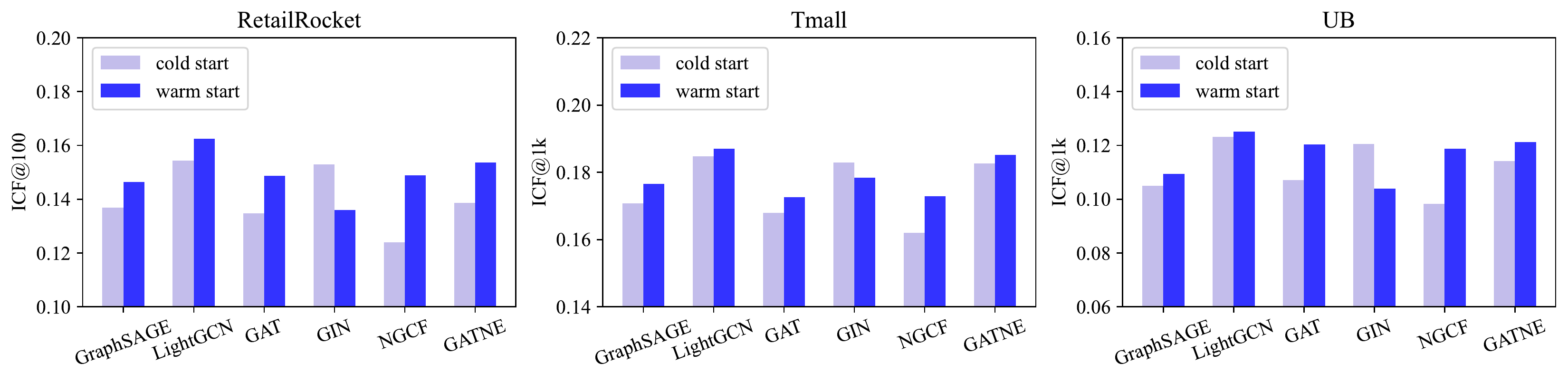}
	\vspace{-0.7cm}
	\caption{The influence of pre-training metapath2vec and warm start for GNN models.}
	\label{fig:warmup}
	
\end{figure*}

In addition, we also find that using the embedding trained by walk-based models to warm start GNN models
can obtain a better performance in less training time compared with pure GNN models.
Figure \ref{fig:warmup} shows that, after warming up from metapath2vec model, 
most GNN models can reach a better performance.
The GNN-based model can aggregate more interaction information from neighbors at one time and generate more informative node representations. With the help of the warm up strategy, the nodes in the graph already have certain structural information, so that the GNN-based model can obtain better aggregated features from the beginning, which helps the training of the GNN model.

\begin{table}[hbt]
	\centering
	\resizebox{0.85\linewidth}{!}{
		\begin{tabular}{lccc}
			\hline
			\textbf{}                 & \textbf{ICF@1k} & \textbf{UCF@1k} & \textbf{U2I@1k} \\ \hline
			\textbf{metapath2vec}     & 0.1654          & 0.1165          & 0.1407          \\
			\textbf{+ Side-info}      & 0.1678          & 0.1164          & 0.1458          \\ \hline
			\textbf{GraphSAGE (mean)} & 0.1708          & 0.1242          & 0.1420          \\
			\textbf{+ Side-info}      & 0.1770          & 0.1227          & 0.1554          \\ \hline
			\textbf{GraphSAGE (sum)}  & 0.1677          & 0.1238          & 0.1496          \\
			\textbf{+ Side-info}      & 0.1761          & 0.1223          & 0.1535          \\ \hline
			\textbf{LightGCN}         & 0.1848          & 0.1404          & 0.1705          \\
			\textbf{+ Side-info}      & 0.1897          & 0.1411          & \textbf{0.1760} \\ \hline
			\textbf{GAT}              & 0.1680          & 0.1230          & 0.1410          \\
			\textbf{+ Side-info}      & 0.1724          & 0.1222          & 0.1512          \\ \hline
			\textbf{GIN}              & 0.1829          & 0.1371          & 0.1593          \\
			\textbf{+ Side-info}      & 0.1895          & 0.1393          & 0.1608          \\ \hline
			\textbf{NGCF}             & 0.1634          & 0.1151          & 0.1366          \\
			\textbf{+ Side-info}      & 0.1722          & 0.1179          & 0.146           \\ \hline
			\textbf{GATNE}            & 0.1826          & 0.1359          & 0.1708          \\
			\textbf{+ Side-info}      & \textbf{0.1904} & \textbf{0.1412} & 0.1606          \\ \hline
		\end{tabular}
	}
	\caption{The influence of side information on various GNN models on Tmall dataset.}
	\label{table:side_info}
	\vspace{-0.2cm}
\end{table}

\subsubsection{GNNs with Side Information (RQ3).} 
\label{label:side_info}
Side information is some discrete features (bag-of features). 
It can be regarded as the node feature so that when meeting a new node (user), 
one can use the node feature embedding to represent it.
To train GNN models with side information, 
we directly sum the features embeddings with the node ID embeddings.
From the Table \ref{table:side_info}, we demonstrate that 
both walk-based and GNN-based models can benefit from adding side information. 
This means that the side information is significant to improve the ability of node embeddings.

From Table \ref{table:side_info},
We discover an interesting phenomenon.
When only node ID embeddings are used, 
LightGCN can outperform other GNN models similar to the finding that transformation and non-linear has negative effects on GNNs \cite{he2020lightgcn}. When appending side information with node ID embeddings,
most of the models can take advantage of it. Moreover, GATNE can surpass the effect of LightGCN with additional features. 

\subsubsection{In-batch Negative Sampling (RQ4).}
\label{label:train_speed}
Typically, distributed training can easily fall into a communication bottleneck, 
which is more likely to be occurred in graph representation learning 
due to the frequent graph sampling.
Therefore, we make some optimizations to the training speed, 
including in-batch negative sampling and ego-graph construction order exchange.

Table \ref{table:negative} shows the impact of in-batch negative sampling. 
We set batch size to 1000 and the number of negative samples to 5 on this experiment. 
For the Rec15 dataset, we train the metapath2vec model with 2 billion edges.
While for Tmall dataset, we use 3 billion edges.
The results show that the training speed of in-batch negative sampling strategy is near 4$\times$ faster than the random negative sampling strategy, while it still maintains the competitive performance.

\subsubsection{Order of Ego Graphs Generation and Pairs Generation (RQ5).}
As described in Section \ref{label:order}, 
to speed up generating training pairs, 
we exchange the order of pairs generation and ego graphs generation, which might change the diversity of training samples with fewer neighborhood sampling operations.
We set the window size $w=2$ and walk length $L=4$.
Table \ref{table:order} shows the influence of the order on LightGCN for Tmall dataset.
We run 150 million pairs for each task and find that
ego graph sampling first can nearly double the speed 
while the performance on three metrics only declines slightly.
Thereby, it is worthwhile to do ego graph sampling before generating pairs from the walk path.

\begin{table}[]
	\setlength\tabcolsep{3pt}
	\resizebox{1.0\linewidth}{!}{
		\begin{tabular}{cccccc}
			\hline
			\textbf{Datasets}               & \textbf{Negatives} & \textbf{Speed (Sec.)} & \textbf{ICF} & \textbf{UCF} & \textbf{U2I} \\ \hline
			\multirow{2}{*}{\textbf{Rec15}} & random             & 5788                     & 0.4477           & 0.4647           & 0.4406           \\
			& in-batch           & 1516                     & 0.4443           & 0.4680            & 0.4415           \\ \hline
			\multirow{2}{*}{\textbf{Tmall}} & random             & 10779                    & 0.1708           & 0.1222           & 0.1434           \\
			& in-batch           & 2675                     & 0.1696           & 0.1255           & 0.1448           \\ \hline
		\end{tabular}
	}
	\caption{The training speed and performance of metapath2vec model 
		between random and in-batch negative sampling.
		We recall the most similar top-100 items for Rec15 and top-1k items for Tmall.}
	\label{table:negative}
\end{table}

\begin{table}[]
	\setlength\tabcolsep{3pt}
	\resizebox{1.0\linewidth}{!}{
		\begin{tabular}{ccccc}
			\hline
			\textbf{Sample Generation Order}                 & \textbf{Speed (Sec.)} & \textbf{ICF} & \textbf{UCF} & \textbf{U2I} \\ \hline
			\textbf{Walk, Pair, Ego} & 10025                    & 0.1392       & 0.1179       & 0.1068       \\
			\textbf{Walk, Ego, Pair} & 6195                     & 0.1341       & 0.1175       & 0.1059       \\ \hline
		\end{tabular}
	}
	\caption{The training speed and performance of LightGCN model.
		We recall the most similar top-1k items on Tmall dataset.}
	\label{table:order}
\end{table}

\subsubsection{Walk-based vs. GNN-based (RQ6).}
\begin{figure}[htbp]
	\centering
	\includegraphics[width=\linewidth]{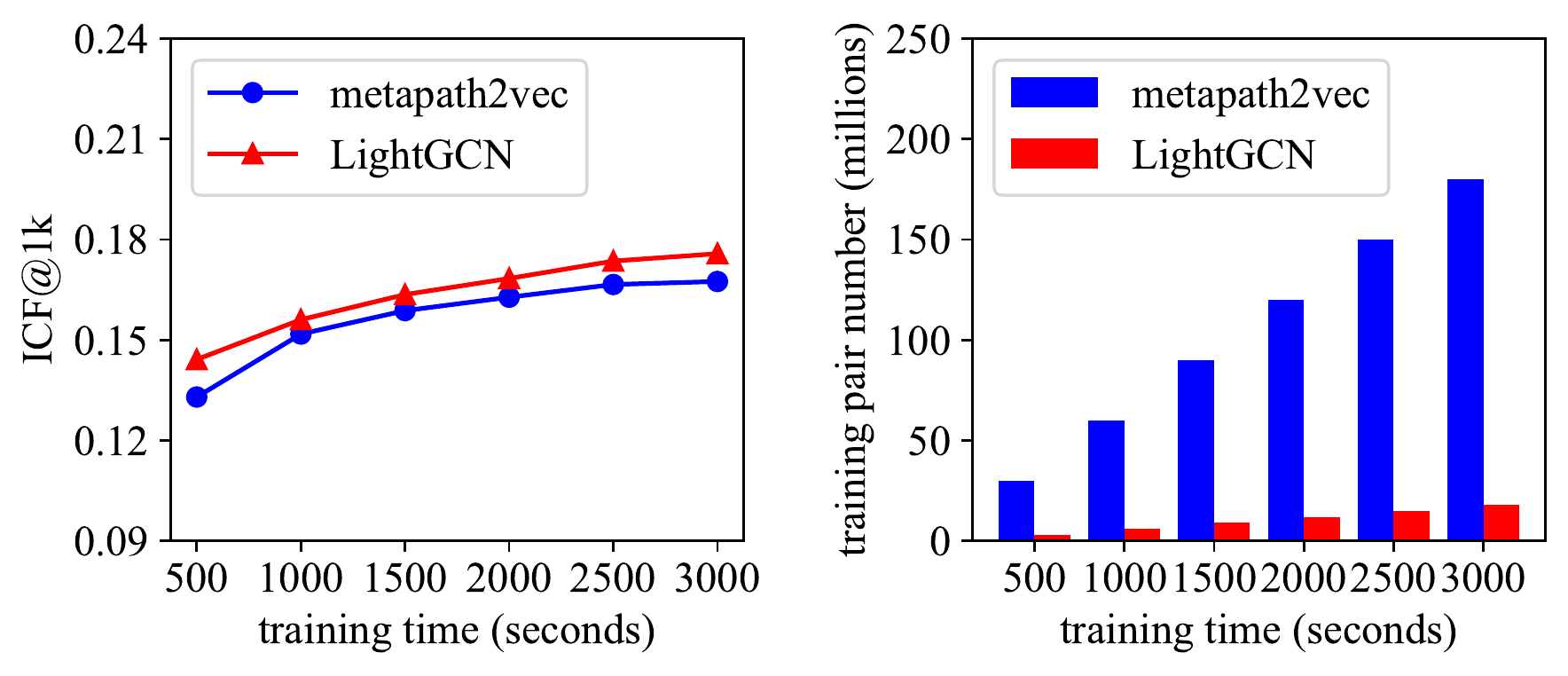}
	\vspace{-0.6cm}
	\caption{The comparison of training speed and convergence 
		with respect to metapath2vec and LightGCN model on Tmall dataset.}
	\label{fig:m2v_vs_gnn}
\end{figure}

In our experiments, 
we find that the training speed of GNN-based models is much slower than walk-based models.
The reason for this is that when training GNN-based models, 
we need to process ego graph sampling and neighbor aggregation
which slows down the mini-batch data generation.
Therefore, we conduct a experiment to see the performance 
of walk-based models and GNN-based models under the same training time.
As shown in Figure \ref{fig:m2v_vs_gnn}, 
the amount of training samples in metapath2vec model is 
nearly 10$\times$ larger than in LightGCN model. 
However, the performance of LightGCN is still better than metapath2vec model.
One possible reason is that GNN-based models can aggregate multiple neighbors at one time, making the convergence faster.

\section{Conclusion}
In this paper, we introduce Graph4Rec, 
a universal toolkit with graph neural networks for recommender systems,
that unifies the paradigm to train GNN models into five components. 
There are three highlights in our Graph4Rec, 
namely large-scale, abundance and easy-to-use.
Extensive experimental results show that 
our Graph4Rec is competitive with other existing systems.
And we also demonstrate some useful experiments 
which may guide our practice in industrial recommender systems.

\bibliographystyle{aaai}
\bibliography{aaai22.bib}

\end{document}